\documentclass{article}

     \usepackage[preprint]{neurips_2020}

\usepackage[utf8]{inputenc} 
\usepackage[T1]{fontenc}    
\usepackage{hyperref}       
\usepackage{url}            
\usepackage{booktabs}       
\usepackage{amsfonts}       
\usepackage{nicefrac}       
\usepackage{microtype}      

\usepackage{graphicx}       
\usepackage{amsmath}        
\usepackage{comment}

\title{Speech Fusion to Face: Bridging the Gap Between Human's Vocal Characteristics and Facial Imaging}

\author{%
  Yeqi Bai$^1$, \ Tao Ma$^2$, \ Lipo Wang$^1$, \ Zhenjie Zhang$^3$\thanks{Corresponding author} \\ 
  $^1$Nanyang Technological University\\
  $^2$Northwestern Polytechnical University\\
  $^3$PVoice Technology\\
  \texttt{ba0001qi@e.ntu.edu.sg, taoma\_nwpu@hotmail.com} \\
  \texttt{elpwang@ntu.edu.sg, zhenjie.zhang@pvoice.io}
}

\begin{document}

\maketitle

\begin{abstract}
While deep learning technologies are now capable of generating realistic images confusing humans, the research efforts are turning to the synthesis of images for more concrete and application-specific purposes. Facial image generation based on vocal characteristics from speech is one of such important yet challenging tasks. It is the key enabler to influential use cases of image generation, especially for business in public security and entertainment. Existing solutions to the problem of \emph{speech2face} renders limited image quality and fails to preserve facial similarity due to the lack of quality dataset for training and appropriate integration of vocal features. In this paper, we investigate these key technical challenges and propose \emph{Speech Fusion to Face}, or SF2F in short, attempting to address the issue of facial image quality and the poor connection between vocal feature domain and modern image generation models. By adopting new strategies on data model and training, we demonstrate dramatic performance boost over state-of-the-art solution, by doubling the recall of individual identity, and lifting the quality score from 15 to 19 based on the mutual information score with VGGFace classifier. 
\end{abstract}

\section{Introduction}
\stepcounter{footnote}
Driven by the explosive growth of deep learning technologies, \cite{choi2018stargan,choi2019stargan,karras2019style}, computer vision algorithms are now capable of generating high-fidelity images, such that humans can hardly distinguish real images from synthesized ones. Researchers in computer vision and other areas are seeking concrete applications to fully unleash the power of image synthesis. One of the most promising applications is the automatic generation of facial images based on the vocal characteristics extracted from human speech \cite{Oh_2019_CVPR,wen2019face}, referred to as \emph{speech2face} in the rest of the paper. It is believed to be the key enabler to new business opportunities in public security, entertainment, and other industries. Specifically, given an audio clip containing the speech from a target individual, the \emph{speech2face} system is expected to visualize the individual's facial image, based on the voice of the target individual \emph{only}.

\begin{figure}[t]
    \centering
    \includegraphics[width=0.7\textwidth]{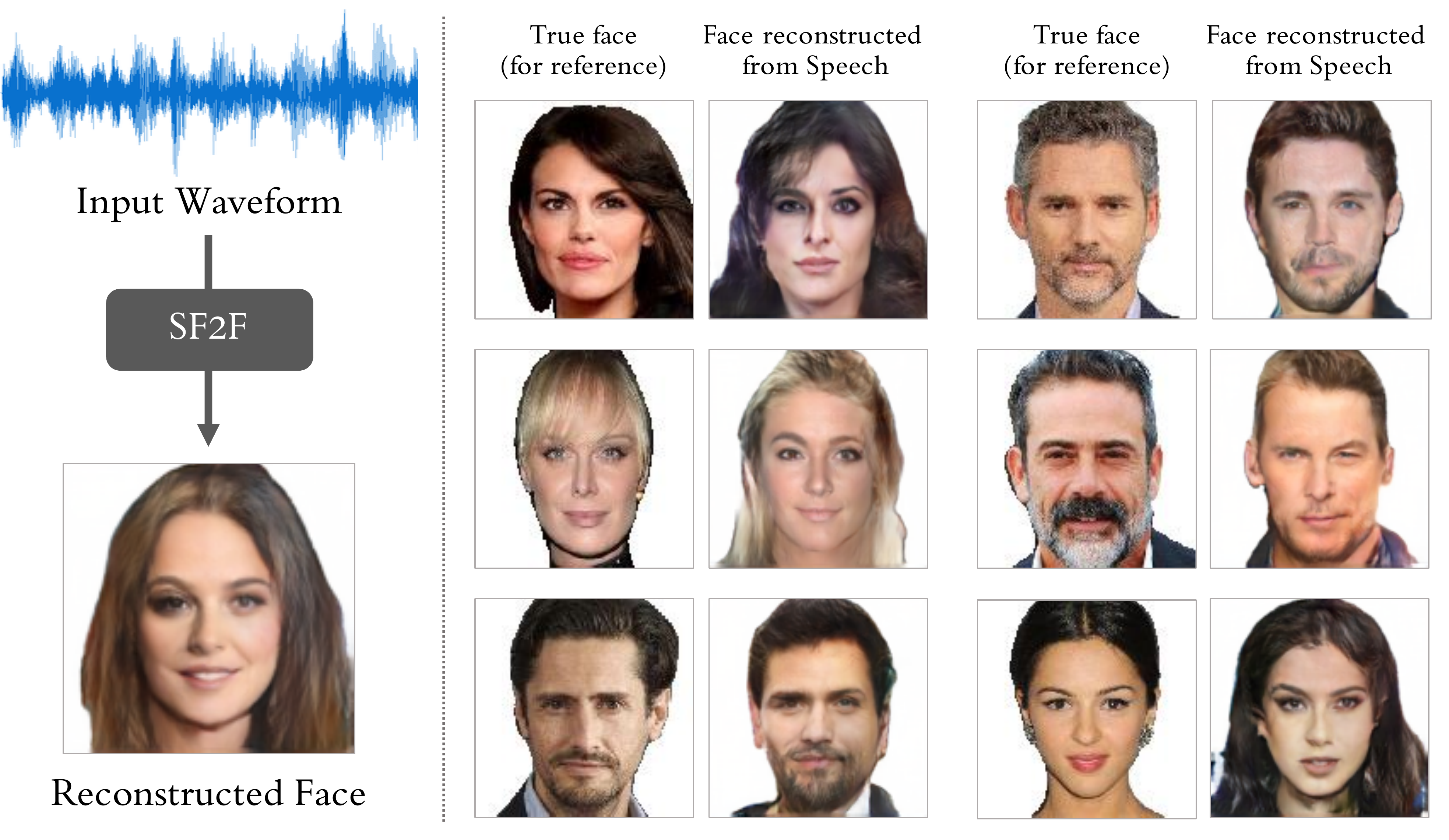}
    \caption{\emph{Left:} The task of reconstructing the face image of an identity from his/her speech waveform. \emph{Right:} A few $128 \times 128$ sample images produced by our SF2F model, where human speech is the only input to the model. The groundtruth faces are shown for reference.}
    \label{fig:overview}
\end{figure}

It is straightforward to understand vocal characteristics reflect important personal features, such as gender and age, which can be easily inferred over an individual's speech. Existing studies have also demonstrated interesting observations on the presence of additional physical features of human face, strongly correlated to the vocal features in the voice. Unfortunately, the quality of the output facial images from these studies remains far from satisfactory, due to the following technical limitations. Firstly, the poor quality of paired dataset of individuals' facial image and clean speech hinders the effective training of generative model for \emph{speech2face}. Secondly, the integration of vocal features into state-of-the-art image synthesis models does not fully utilize the rich information from the speech of the individuals.

To address these technical challenges, we propose \emph{speech fusion to face}, or SF2F in short, in this paper. Fig. \ref{fig:overview} shows sample results of our method. We attempt to improve on top of the existing approaches in two general directions. To tackle the problem of poor data quality, we enhance the VoxCeleb dataset \cite{chung2018voxceleb2,nagrani2017voxceleb} by filtering and rebuilding the face images of the celebrities. In Fig. \ref{fig:quality_comparison}, we present example images from filtered VGGFace dataset used in \cite{parkhi2015deep} and our refined face image dataset. Common problems associated with the original face images from VGGFace dataset include (1) plenty of monolithic photos containing limited details of the faces; (2) the shooting angle and facial expression are highly diverse and technically difficult to normalize; (3) the background of the face image is not appropriately eliminated. Given these problems, we target to rebuild a high-quality face database for individuals in VoxCeleb database. By carefully retrieving images of these celebrities from the Internet and filtering with specific conditions, we generate HQ-VoxCeleb database with example figures shown in Fig. \ref{fig:quality_comparison}, with highly normalized facial images on different aspects.

Existing studies on speaker recognition, e.g., \cite{wan2018generalized}, imply that vocal characteristics are highly unstable in the speech. Speaker recognition/verification algorithms usually extract short-time vocal features from speech and combine these features for final decision making. Such strategies cannot be directly copied to the solution for \emph{speech2face} system because simple aggregation may lose some key information only available in certain pronunciations. In order to exploit the short-term features for face image generation, we introduce a new fusion strategy over speech extraction algorithms, which applies merging over the face images with different embeddings instead of the original embeddings. A new training scheme, with a new combination of loss functions, is introduced accordingly in order to maximize the information flowing from speech domain to facial imaging domain.

%
The key contributions of the paper are summarized as follows: (1) we introduce an enhanced face image dataset based on VoxCeleb, containing over 3,000 individual identities with high-quality front face images; (2) we present a new fusion strategy over the short-term vocal characteristics to stabilize the facial features for the synthetic image generation; (3) we propose a new loss function for the neural generative model in order to better preserve the vocal features in the facial generation process; (4) we evaluate the performance of the proposed model on both image fidelity and facial similarity to ground-truth and validates the performance gain over the state-of-the-art solutions.


\section{Related Work}\label{sec:related}

\noindent\textbf{Generative Models.}
Plenty of generative frameworks have been proposed in recent years. {Auto-regressive approaches} such as PixelCNN and PixelRNN \cite{oord2016pixel} generate images pixel by pixel, modeling the conditional probability distribution of sequences of pixels. {Variational Autoencoders} (VAEs) \cite{kingma2013auto,NIPS2016_6528} jointly optimize a pair of encoder and decoder. The encoder transforms the input into a latent distribution, and the decoder synthesizes images based on the latent distribution. Remarkably, {Generative Adversarial Networks (GANs)} \cite{NIPS2014_5423,radford2015unsupervised} which adversarially train a pair of generator and discriminator, have achieved the best visual quality.

\noindent\textbf{Visual generation from audio.}
The generation of visual information from various types of audio signals has been studied extensively. There exist approaches \cite{taylor2017deep,karras2017audio,suwajanakorn2017synthesizing} to generate an animation that synchronizes to input speech by mappings from phoneme label input sequences to mouth movements. Regarding pixel-level generation, Sadoughi and Busso's approach \cite{sadoughi2019speech} synthesizes talking lips from speech, and X2Face model \cite{wiles2018x2face} manipulates the pose and expression of conditioning on an input face and audio. Wav2Pix model \cite{giro2019wav2pix} is most relevant to our work, in the sense that it generates face images from speech, with the help of adversarial training. Their objective is to reconstruct the face texture of the speaker including expression and pose. In this paper, our objective is different, targeting to reconstruct a frontal face with a neutral expression, such that most facial attributes of the groundtruth are preserved.

\noindent\textbf{Face-speech association learning.}
The associations between faces and speech have been widely studied in recent years. Cross-modal matching methods by classification \cite{shon2019noise,nagrani2018learnable,nagrani2018seeing,wen2018disjoint} and metric learning \cite{kim2018learning,horiguchi2018face} are adopted in identity verification and retrieval. Cross-modal features extracted from faces and speech are applied to disambiguate voiced and unvoiced consonants \cite{chung2017lip,ngiam2011multimodal}; to track active speakers of a video \cite{hoover2017putting,gebru2015tracking}; or to predict emotion \cite{albanie2018emotion} and lip motions \cite{ngiam2011multimodal,andrew2013deep} from speech.

\noindent\textbf{Face reconstruction from speech.}
\emph{Speech2face} is an emerging topic in computer vision and machine learning, aiming to reconstruct face images from a voice signal based on existing sample pairs of speech and face images. Oh \textit{et al.} \cite{Oh_2019_CVPR} propose to use a voice encoder that predicts the face recognition embedding, and decode the face image using a pre-trained face decoder \cite{Cole_2017_CVPR}. As the most similar approach to ours, Wen \textit{et al.} \cite{wen2019face} employ an end-to-end encoder-decoder model together with GAN framework, based on filtered VGGFace dataset \cite{parkhi2015deep}. In this paper, we propose a suite of new approaches and strategies to improve the quality of face images on top of these studies.

\section{Data Quality Enhancement}\label{section:dataset}



\begin{figure}[t]
    \centering
    \includegraphics[width=0.7\textwidth]{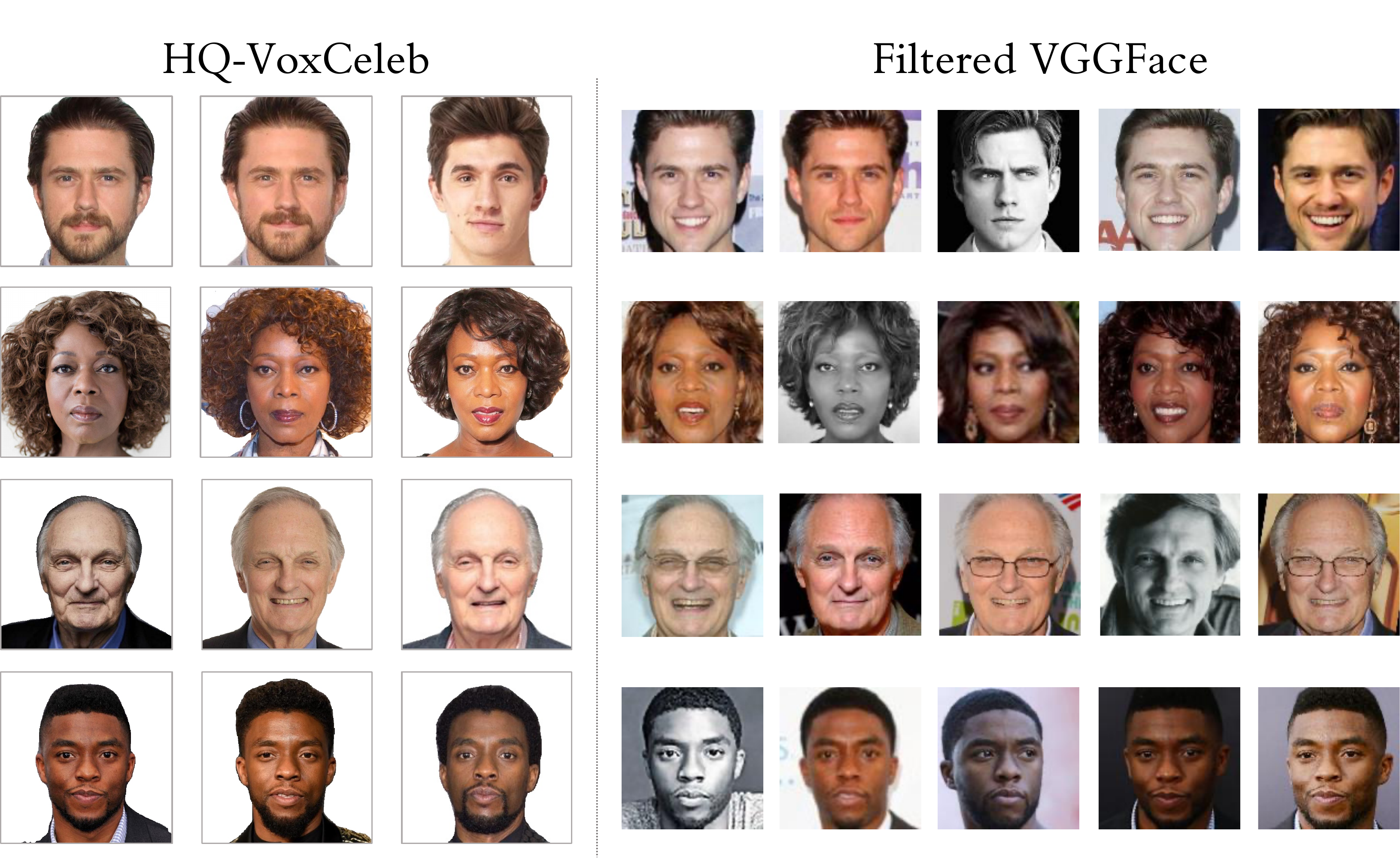}
    \caption{The quality variance of the original manually filtered VGGFace dataset (on the right of the figure) \cite{parkhi2015deep} diverts the efforts of face generator model to the normalization of the faces. By filtering and correcting the face database (on the left), the machine learning models could focus more on linking facial and vocal features.}
    \label{fig:quality_comparison}
\end{figure}

As demonstrated in Fig. \ref{fig:quality_comparison}, the poor quality of training dataset for \emph{speech2face} is one of the major factors hindering the improvement of \emph{speech2face} performance. To eliminate the negative impact of the training dataset, we carefully design and build a new high-quality face database on top of VoxCeleb dataset \cite{chung2018voxceleb2,nagrani2017voxceleb}, such that face images associated with the celebrities are all at reasonable quality. To fulfill the vision of data quality, we set a number of general guidelines over the face images as the underlying measurement on quality, covering attributes including face angles, lighting conditions, facial expressions, and image background.
In order to fully meet the standards as listed above, we adopt a data processing pipeline to build the enhanced \emph{HQ-VoxCeleb} dataset for our \emph{speech2face} model training, with details available in Appendix \ref{sec:hq-vox}. In Table \ref{tab:table1}, we summarize the statistics of the result dataset after the adoption of the processing steps above. In Appendix \ref{sec:hq-vox}, we compare HQ-VoxCeleb with existing audiovisual datasets, to justify HQ-VoxCeleb is the most suitable dataset for end-to-end learning of \emph{speech2face} algorithms. HQ-VoxCeleb dataset will be released later.


\vspace{-0mm}
\begin{table}[h!]
  \begin{center}
    \caption{Statistics of HQ-VoxCeleb dataset, where speech data is acquired from the identity intersection with VoxCeleb dataset.}
    \label{tab:table1}
    \begin{tabular}{c|c|c|c|c}
      \toprule
      \textbf{ Attribute } & \textbf{ Total } & \textbf{ Train. Set } & \textbf{ Val. Set } & \textbf{ Test Set } \\
      \midrule
      \#Identity & 3,638 & 2,890 & 370 & 378 \\
      \#Image & 8,028 & 6,375 & 859 & 794 \\
      \#Speech & { 609,700 } & 478,009 & 63,337 & 68,354 \\
      { Avg. Reso. } & { 505.41$\times$505.41} & { 505.78$\times$505.78 } & { 503.63$\times$503.63 }  & { 504.37$\times$504.37 } \\
      \bottomrule
    \end{tabular}
  \end{center}
\end{table}

\section{Model and Approaches}\label{sec:approach}

\begin{figure}[t]
\centering
\includegraphics[width=.8\columnwidth]{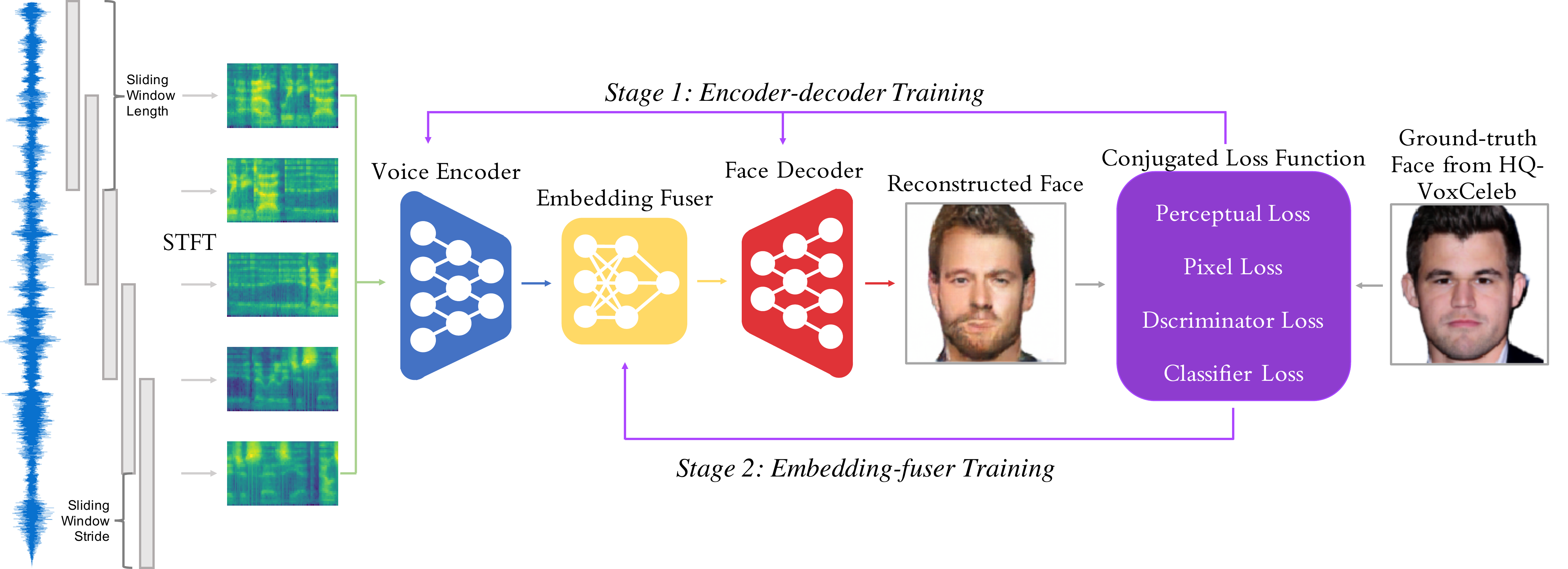}
\caption{Overview of the framework of our proposed SF2F
}
\label{fig:framework}
\end{figure}

The overall architecture of our proposed SF2F model is illustrated in Fig. \ref{fig:framework}. 
%
The pipeline of our model follows the popular encoder-decoder architecture used in most of the generative models. We refine the structure to make a better fit for the \emph{speech2face} task. At inference, we apply a rolling window over the input speech audio. Each window is regarded as an individual local speech segment and the corresponding features from one segment are extracted and stored in a 1D vector by a speech encoder. The global facial features are generated by applying an attention fuser over the sequence of local facial features from the sliding windows. 
%
%
As shown in Fig. \ref{fig:framework}, the whole training procedure consists of two stages. In the first stage, the voice encoder and face decoder are first optimized to correlate facial features with vocal features extracted from short-time audio windows ranging from 1 second to 1.5 seconds and reconstructing faces based on the features. In the second stage, longer range of audio (global audio) ranging from 6 seconds to 25 seconds are converted into a series of local audio segments with a sliding window; the voice encoder converts the audio segments into corresponding embeddings. An embedding fuser follows to obtain a global facial embedding conditioned on the sequence of local facial embeddings. The weights of voice encoder and face decoder remain fixed in the second stage, while the two training stages share the same conjugated loss function. 
%
%
%
On the other hand, our approach enables the model to save efforts on learning useless variations mostly contained in long audios, such as the prosody of the speaker and linguistic features of the transcript, and consequently prevents the model from over-fitting. 
Moreover, the proposed framework maximizes the information extraction from short audios containing most important vocal features with matching features in the facial imaging domain. 
%

\noindent\textbf{Pre-Processing.} Over every speech audio waveform, we apply a sliding window at fixed width of 1.25 seconds with 50\% overlap. Short-time Fourier Transform (STFT) \cite{muller2015fundamentals} computes the mel-spectrogram of each window. The input waveform is thus converted into a sequence of mel-spectrograms $M =  \left\{m_{i} \in \mathbb{R} ^ {{T} \times {F}} \right\}$, where $T$ denotes the number of frames in the time domain, and $F$ denotes frequency. As we use an end-to-end training pipeline, instead of relying on pre-trained face decoder as in \cite{Oh_2019_CVPR}, we expect the groundtruth face images to contain least irrelevant variations, including pose, lighting, and expression. This is enforced by our data enhancement scheme in Section \ref{section:dataset}. Specifically, each groundtruth face image is denoted by $I \in \mathbb{R} ^ {3\times{H}\times{W}}$ in the rest of the paper. 

\noindent\textbf{Voice Encoder.}
We use a 1D CNN composed of several 1D Inception modules \cite{szegedy2015going} to process mel-spectrogram. The voice encoder module converts each short input speech segment into a predicted facial feature embedding. The architecture of the 1D Inception module and the voice encoder is described with details in Appendix \ref{sec:model}. The encoder's basic building block is a 1D Inception module, where each Inception module consists of four parallel 1D CNNs, each with kernel size ranging from 2 to 7 and stride 2 with convolution along the time domain, followed by ReLU activation and batch normalization \cite{ioffe2015batch}. The Inception module models various ranges of short-term mel-spectrogram dependency, and enables more accurate information flow from voice domain to image domain, compared to plain single-kernel-size CNN used in \cite{wen2019face}. 
The number of channels grows from lower layers to higher layers in the stack, in order to reflect the bandwidth of the frequency domain to the corresponding layers. On top of the stack of basic building blocks, an average pooling layer outputs a predicted facial embedding $e_{i} \in \mathbb{R} ^ d$ for each mel-spectrogram segment $m_i$. The embedding $e_i$ is normalized as $e_i=\frac{e_i}{\max(\lVert e_i \rVert_2, \epsilon)}$, in which $\epsilon = 10^{-12}$ by default.


\noindent\textbf{Face Decoder.} The decoder reconstructs target individual's face image based on embeddings from the individual's speech segments. Since HQ-VoxCeleb dataset eliminates unnecessary and redundant variance on irrelevant features, the face decoder is expected to target the most distinctive features of faces. The basic building blocks of our face decoder include a 2D bilinear upsampling operator, a 4$\times$4 convolution operator, a ReLU activation, and a batch normalization. On top of stacked basic building blocks, a 1$\times$1 convolution with linear activation is deployed to generate the output $I'$.

\begin{figure}[t]
\centering
\includegraphics[height=5.5cm]{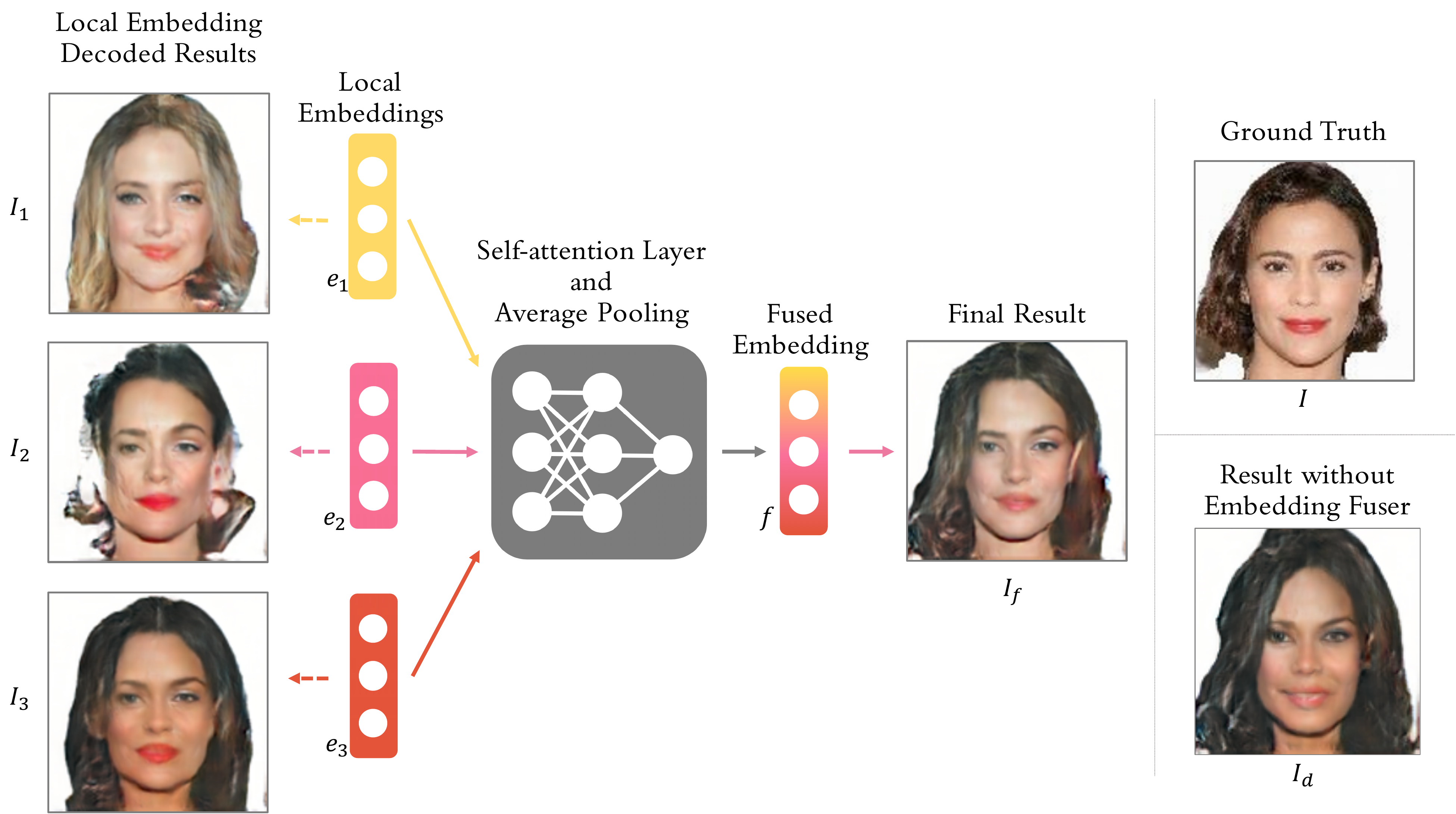}
\caption{Demonstration of the functionality of embedding fuser. In this case, three local embeddings $\{e_1, e_2, e_3\}$ are fused into global embedding $f$. Faces $\{I_1, I_2\}$ decoded from embeddings $\{e_1, e_2\}$ contains obvious visual defects due to the noises in the corresponding audio segments. Face $I_3$ renders better image quality, while the face shape and skin color is not even close to the ground truth. By fusing $\{e_1, e_2, e_3\}$ into $f$, the face decoder outputs $I_f$ combining the merits of $\{I_1, I_2, I_3\}$ and avoid the local defects within $\{I_1, I_2, I_3\}$. Without embedding fuser, the result image can hardly distinguish the appropriate features for the face generation procedure.}
\label{fig:fuser}
\end{figure}

\noindent\textbf{Embedding Fuser.}
The most important part of our model is the fuser of the embeddings from the speech segments. The rationale behind the independent extraction of local features from the speech segment is to enable the model to focus on the vocal characteristics of the individual. When the voice encoder is directly applied to long speech audio, for example, the output features may contain features over the texts in the speech, as well as prosodic features on the emotion and presentation style of the speakers. These features are speaker-independent and therefore irrelevant to the features of the individuals in facial imaging domain. However, embeddings from speech segments are not stable by themselves. Because of the variance of text content in the speech segment, each embedding may reflect completely different vocal characteristics useful to the \emph{speech2face} task. The embedding fuser is illustrated in the example in Fig. \ref{fig:fuser}.



Given a sequence of mel-spectrograms $M = \{m_1, m_2, ..., m_{T-1}, m_T\}$ generated Short-Time Fourier Transform (STFT) over $T$ consecutive speech segments, voice encoder transforms $M$ into a sequence of embedding vectors $E_{1:T} = \{e_1, e_2, ..., e_{T-1}, e_T\}$, each of which contains certain vocal features correlated to different facial features. In order to preserve the useful information from all of the embeddings, embedding fuser first processes $E_{1:T}$ with a self-attention layer. Given a voice embedding vector $e_i$, its attention score to another voice embedding vector $e_j$ is calculated by $s(e_i, e_j) = e_i ^T W_a e_j$.
Based on the scores between the embedding vectors, $\beta_{i, j}$ indicates the degree of $e_i$ attending to $e_j$, measured by $\beta_{i, j} = \frac{\exp(s(e_i, e_j))}{\sum {_{j=1} ^N}\exp(s(e_i, e_j))}$.
The output of $e_i$, for each $1\leq i\leq T$, attending to the whole sequence $E$ is finalized as	$a_i = \sum {_{j=1}^T} \beta _{i, j} e_j$.
Here, attention output $a_i$ is concatenated with voice embedding vector $e_i$ and then goes through a linear transformation, generating the output of a fine-grained feature $f_i = W_f [a_i, e_i] + b_f$.
Finally, average pooling is performed along the time dimension, $f = \frac{1}{T} \sum {_{i=1}^T} f_i$, where $f$ is the globally fused embedding vector, with dimensionality identical to $e_i$ and $f_i$. 


\noindent\textbf{Discriminators.}
To improve the generative capability of our proposal, two discriminators, namely $D_{real}$ and $D_{id}$, are employed as the adversary to the generative model. 
The discriminator $D_{real}$ is the traditional binary discriminator as introduced in standard Generative Adversarial Network (GAN) \cite{NIPS2014_5423} that distinguishes the synthesized images from the real images, by optimizing over the following objective function: $\mathcal{L}_{G} = \mathop{\mathbb{E}} \limits_{x \sim p_{\mathrm{real}}} \log D_{real}(x) + \mathop{\mathbb{E}} \limits_{x \sim p_{\mathrm{fake}}} \log (1 - D_{real}(x))$, where $x \sim p_{\mathrm{real}}$ and $x \sim p_{\mathrm{fake}}$ represent the distributions of real and synthesized images, respectively. 
%
Instead of telling difference between real and fake images, the discriminator $D_{id}$ attempts to classify the images into the identity of the individual in the image. A standard identity classifier aims to optimize the following objective function: $\mathcal{L}_{id} = \mathop{\mathbb{E}} \limits_{x \sim p_{\mathrm{real}}} \sum_{i=1}^{C} y_i \log(D_{id}(x)_i) $.

Conditioned on human speech, the generator network is expected to synthesize human face images, which can be correctly classified by $D_{id}$. This is achieved by minimizing the objective function $
\mathcal{L}_{C} = \mathop{\mathbb{E}} \limits_{x \sim p_{\mathrm{fake}}} \sum_{i=1}^{C} - y_i \log(D_{id}(x)_i)$ on generated images .

\noindent\textbf{Training Scheme.}
%
As shown in Fig. \ref{fig:framework}, the weights associated with the encoder-decoder and the embedding fuser are updated in two stages, respectively.
The unified conjugated loss function consists of four different types of losses:

(1) \emph{Image Reconstruction Loss.} $\mathcal{L}{_1} = \lVert I - \hat I \rVert {_1} $ penalizes the $L_1$ differences between the ground-truth image $I$ and the reconstructed image $\hat I$.
%
%
(2) \emph{Adversarial Loss.} Image adversarial loss $\mathcal{L}_{G}$ from $D_{real}$ encourages generated face image to appear photo-realistic; (3) \emph{Auxiliary Classifier Loss.} $\mathcal{L}_{C}$ ensures generated objects to be recognizable and well-classified by $D_{id}$; (4) \emph{FaceNet Perceptual Loss.} Image perceptual loss $\mathcal{L}_P$ penalizes the $L_1$ difference in the global feature space between the ground truth image $I$ and the reconstructed image $\hat I$, while object perceptual loss. Inspired by \cite{johnson2016perceptual}, 
%
%
we add a lightweight perceptual loss between generated images and ground truth images to keep the perceptual similarity. This loss not only improves the quality of the generated images, but also enhances the similarity of the output images to the ground truth faces.
The loss function of our model is finally formulated as, $\mathcal{L}_{conj} = \lambda_1 \mathcal{L}{_1} + \lambda_2 \mathcal{L}_{G} + \lambda_3 \mathcal{L}_{C} + \lambda_4 \mathcal{L}_P$, where $\{\lambda_1,\lambda_2,\lambda_3,\lambda_4\}$ are the hyper-parameters used to balance the loss functions.

\section{Experiments}\label{sec:experiment}

%

\noindent\textbf{Implementation Details.} We train all our SF2F models using Adam optimizer \cite{dpk2015adam} with learning rate 5e-4 and batch size 256; the encoder-decoder training takes 120,000 iterations, and the fuser training takes 360 iterations. $\lambda_1$, $\lambda_2$, $\lambda_3$, $\lambda_4$ are set at 10, 1, 0.05 and 100, respectively. We implement discriminator with ReLU activation and batch normalization. More information of the dataset is covered in Section \ref{section:dataset}.
%


\noindent\textbf{Metric on Similarity.} We evaluate the quality of synthesis outputs by measuring the perceptual similarity between the reconstructed face and its corresponding groundtruth \cite{Oh_2019_CVPR}. 
By feeding the generated face image to a pre-trained FaceNet \cite{schroff2015facenet}, we generate the face embedding by retrieving the output of the last layer prior to softmax. Given the embeddings from the groundtruth, denoted by $U=\{u_1,u_2,\ldots,u_N\}$, and the embeddings from the synthesized images, denoted by $U'=\{u'_1,u'_2,\ldots,u'_N\}$, over $N$ individuals in our test dataset, we calculate and report the average cosine similarity, i.e., $\sum^N_{i=1}\cos (u_i,u'_i)/N$, and the average $L_1$ distance, i.e., $\sum^N_{i=1} \|u_i-u'_i\|/N$.

\noindent\textbf{Metric on Retrieval Performance.} We also validate the usefulness of the output face images by querying the test face dataset with the generated face as the reference image, aiming to retrieve the identity of the speaker \cite{Oh_2019_CVPR}. Specifically, Recall@K with $K=1,2,5,10$ are reported, which is the ratio of query face images with its groundtruth identity included in the top-$K$ similar faces.


\noindent\textbf{Image Quality Evaluation.} Existing studies \cite{Oh_2019_CVPR,wen2019face} do not report the general quality of the generated face images, partly because of the missing of appropriate metric on face image synthesis quality. While Inception Score \cite{salimans2016improved} is commonly employed in studies of image generation \cite{johnson2018image,xu2018attngan,yikang2019pastegan,zhang2017stackgan}, it is designed to cover a wide variety of object types in the images. We introduce and report a new quality metric based on VGGFace classifier, which is calculated in a similar way as Inception Score. Inception classifier is simply replaced with VGGFace classifier. The mutual information based on the probabilities over the identities, i.e., $\mathbf{VFS}(G) = \exp \bigg( \frac{1}{N} \sum_{i=1}^N D_{KL} \Big( p(y|\mathbf{x}^{(i)}) || p(y) \Big) \bigg)$, is used to measure image quality. Here, $\mathbf{x}^{(i)}$ stands for the $i$-th image synthesized by the generator $G$, $y$ is the prediction by VGGFace, and $D_{KL}$ denotes KL Divergence.
%
%
%
To justify the adoption of VGGFace Score, we compare it against Inception Score over the \emph{real} images in HQ-VoxCeleb dataset, at different resolutions. The results in table \ref{tab:vfs} show that Inception Scores are locked in a small range between 2.0 and 2.6 regardless of the image resolution, implying the total information of the \emph{real} faces could be encoded with 1 bit only! This is because Inception V3 is not designed for face classification and thus skips the variance of the details on the faces. VGGFace Score, on the other hand, demonstrates a much more reasonable evaluation of face image quality. 



\begin{table}[h]
  \vspace{0mm}
  \begin{center}
    \caption{VGGFace and Inception Scores on face images at different resolutions in HQ-VoxCeleb}
    \label{tab:vfs}
    \small
    \begin{tabular}{c|c|c|c}
      \toprule
      \textbf{Resolution} & \textbf{64 $\times$ 64} & \textbf{128 $\times$ 128} & \textbf{256 $\times$ 256}\\
      \midrule
      \textbf{Inception Score} & {2.074 $\pm$ 0.067} & {2.286 $\pm$ 0.036} & {2.583 $\pm$ 0.108}\\
      \midrule
      \textbf{VGGFace Score} & {48.720 $\pm$ 1.688} & {51.048 $\pm$ 3.633} & {50.525 $\pm$ 1.862}\\
      \bottomrule
    \end{tabular}
  \end{center}
\end{table}


\noindent\textbf{Comparison with existing work.} We include voice2face \cite{wen2019face} as the state-of-the-art baseline in this group of experiments. To make a fair comparison, we implement voice2face by ourselves and train its model with our {HQ-VoxCeleb} dataset. Both voice2face and our SF2F are trained to generate images with 64$\times$64 pixels, and evaluated with 1.25 seconds and 5 seconds of human speech clips. As is presented in Table \ref{table:64_performace_comparison}, SF2F outperforms voice2face by a large margin on all metrics. By deploying fuser in SF2F, the maximal recall@10 reaches 36.65\%, almost doubling the recall@10 of voice2face at 20.82. Performance comparison on filtered VGGFace dataset is discussed in Appendix \ref{sec:fvg_perf}.


\begin{table}[h]
  \setlength{\tabcolsep}{3pt}
  \caption{Performance on HQ-VoxCeleb in feature similarity, retrieval recall@K, and VGGFace Score}
  \small
  \begin{center}
  \begin{tabular}{cc|cc|cccc|c}
    \toprule
    \multicolumn{2}{c}{ \textbf{Setting} } & \multicolumn{2}{|c}{\textbf{ Similarity }} & \multicolumn{4}{|c}{\textbf{ Retrieval }} & \multicolumn{1}{|c}{\textbf{ Quality }} \\ 
    {Method} & {Len.} & {cosine} & {L1} & {R@1} & {R@2} & {R@5} & {R@10} & {VFS} \\
    \midrule
    {random} & {-} & {-} & {-} & {1.00} & {2.00} & {5.00} & {10.00} & {-}\\
    \midrule
    {voice2face} & {1.25 sec} & {0.202} & {23.46} & {1.77} & {3.55} & {14.60} & {20.33} & {15.47 $\pm$ 0.67} \\
    {SF2F (no fuser)} & {1.25 sec} & \textbf{0.304} & \textbf{19.22} & \textbf{5.32} & \textbf{8.10} & \textbf{19.27} & \textbf{32.80} & \textbf{18.59 $\pm$ 0.87} \\
       \midrule
   	{voice2face} & {5 sec} & {0.205} & {23.17} & {1.64} & {3.64} & {15.12} & {20.82} & {15.68 $\pm$ 0.64} \\
   	{SF2F (no fuser)} & {5 sec} & {0.294} & {18.51} & {4.11} & {8.38} & {19.43} & {33.97} & {18.54 $\pm$ 1.02} \\
   	{SF2F} & {5 sec} & \textbf{0.317} & \textbf{17.91} & \textbf{5.75} & \textbf{9.32} & \textbf{20.36} & \textbf{36.65} & \textbf{19.49 $\pm$ 0.59} \\
    \bottomrule
  \end{tabular}
  \end{center} 
  \label{table:64_performace_comparison}
\end{table}

\noindent\textbf{Effect of input audio length.} We examine different audio length for the extraction of vocal features. It is crucial to capture the most important vocal features while reserving minimal irrelevant linguistic and prosody information from speech. 
In Table \ref{tab:time_comparison}, the results show speech audios between 1 second and 1.5 seconds achieve the best performance on most of the feature similarity and retrieval recall metrics. This justifies the assumption that only short audio's vocal characteristics are useful to the face reconstruction task. VFS remains stable with different audio lengths, implying the capacity of the information flow from vocal domain to facial domain is fairly static.


 \begin{table}[h]
  \setlength{\tabcolsep}{3pt}
  \caption{Performance of SF2F (no fuser) on HQ-VoxCeleb, with audios limited to different lengths.}
  \small
  \begin{center}
  \begin{tabular}{c|cc|cccc|c}
    \toprule
    \multicolumn{1}{c}{ \textbf{Setting} } & \multicolumn{2}{|c}{\textbf{ Similarity }} & \multicolumn{4}{|c}{\textbf{ Retrieval }} & \multicolumn{1}{|c}{\textbf{ Quality }} \\ 
    {Audio Length} & {cosine} & {L1} & {R@1} & {R@2} & {R@5} & {R@10} & {VFS} \\    
    \midrule
    {0.5 - 1 sec} & {0.230} & {21.85} & {4.67} & {8.38} & {15.40} & {26.05} & {18.17 $\pm$ 0.75} \\
    {1 - 1.5 sec} & \textbf{0.296} & \textbf{18.51} & \textbf{5.02} & {8.60} & \textbf{19.43} & \textbf{32.25} & {18.53 $\pm$ 0.72} \\
    {1.5 - 3 sec} & {0.294} & {18.79} & {4.96} & \textbf{8.78} & {18.59} & {29.88} & \textbf{18.67 $\pm$ 1.72} \\
    {3 - 6 sec} & {0.274} & {19.21} & {4.33} & {8.43} & {17.10} & {30.14} & {18.05 $\pm$ 1.14} \\
    {6 - 10 sec} & {0.227} & {21.92} & {3.34} & {6.58} & {14.73} & {26.92} & {18.40 $\pm$ 2.03} \\

    \bottomrule
  \end{tabular}
  \end{center} 
  \label{tab:time_comparison}
\end{table}


\begin{figure}[t]
\centering
\includegraphics[width=0.9\columnwidth]{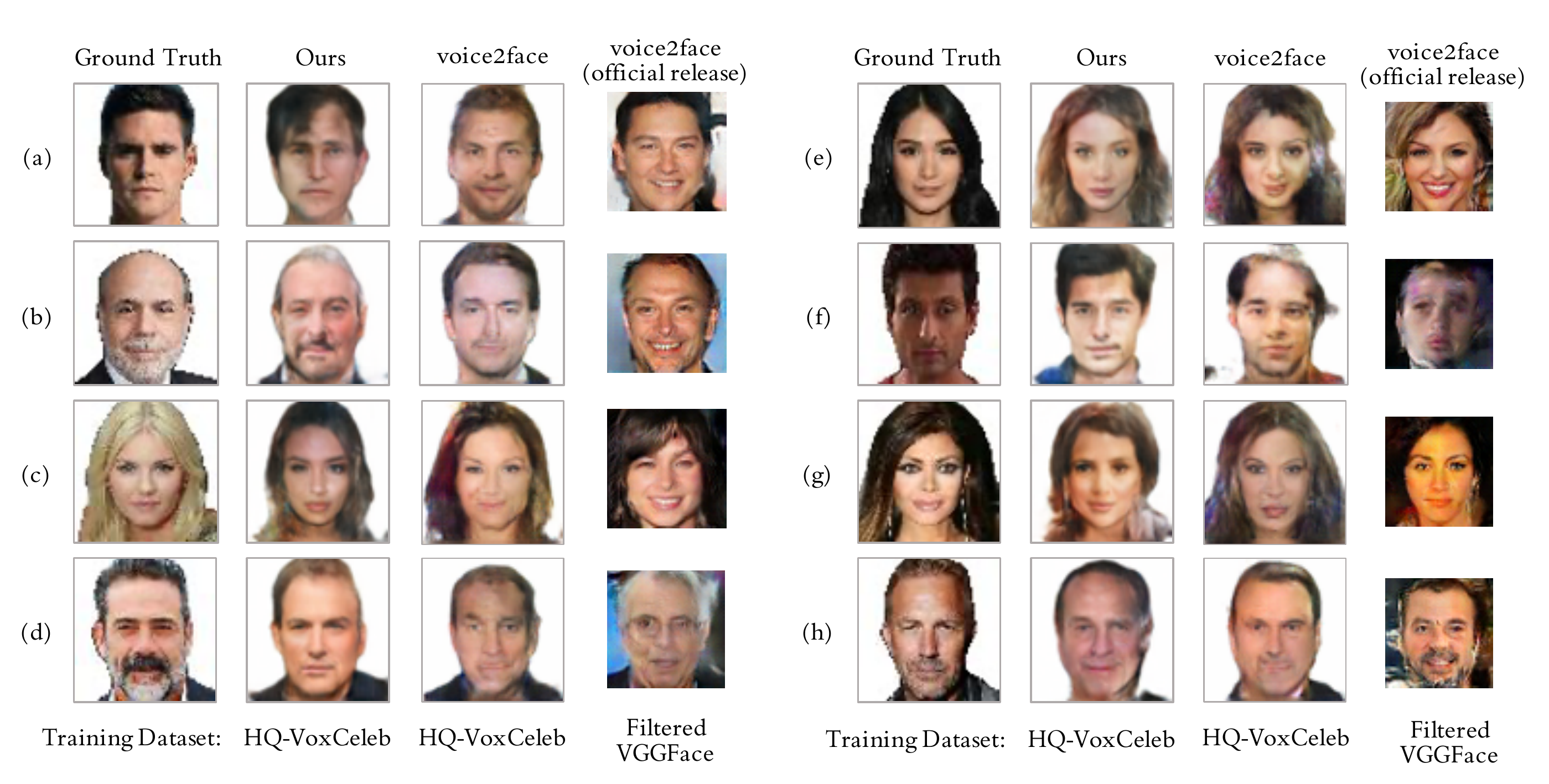}
\caption{Examples of 64 $\times$ 64 generated images using SF2F and voice2face}
\label{fig:qualitative}
\end{figure}

\noindent\textbf{Effect of output resolution.} SF2F is also trained to generate large images with  128$\times$128 pixels. Two different approaches are tested over higher resolution, including a single-resolution SF2F model optimized to generate 128$\times$128 pixels only, and a multi-resolution model optimized to generate both 64$\times$64 pixels and 128$\times$128 pixels. Both of the solutions are equipped with the same conjugated loss function. Table \ref{tab:reso_comparison} compares the performance of SF2F with fusers trained under different schemes with speech length at 5 seconds. Single resolution model performs worse than the other two, suffering from the unstable optimization gradient over a deeper face decoder. Multi-resolution model achieves comparable performance to single-resolution model. This is because most of the speaker-dependent facial features are already included in 64$\times$64 pixels. The addition of more details with higher resolution does not benefit at all.

\begin{table}[h]
  \setlength{\tabcolsep}{3pt}
  \vspace{0mm}
  \caption{Performance of SF2F on HQ-VoxCeleb, with image data resized to different resolution}
  \small
  \begin{center}
  \begin{tabular}{cc|cc|cccc|c}
    \toprule
    \multicolumn{2}{c}{ \textbf{Setting} } & \multicolumn{2}{|c}{\textbf{ Similarity }} & \multicolumn{4}{|c}{\textbf{ Retrieval }} & \multicolumn{1}{|c}{\textbf{ Quality }} \\ 
    {Optimization} & {Resolution} & {cosine} & {L1} & {R@1} & {R@2} & {R@5} & {R@10} & {VFS} \\    
    \midrule
    {single-reso.} & {64 $\times$ 64} & \textbf{0.317} & {17.91} & {5.75} & \textbf{9.32} & \textbf{20.36} & \textbf{36.65} & {19.49 $\pm$ 0.59} \\
    {single-reso.} & {128 $\times$ 128} & {0.278} & {18.54} & {5.02} & {8.60} & {19.43} & {32.25} & {18.53 $\pm$ 1.75} \\
    {multi-reso.} & {128 $\times$ 128} & {0.313} & \textbf{17.46} & \textbf{6.10} & {9.25} & {18.77} & {35.38} & \textbf{20.10 $\pm$ 0.47} \\

    \bottomrule
  \end{tabular}
  \end{center} 
  \label{tab:reso_comparison}
\end{table}

\noindent\textbf{Qualitative results.} We also directly compare the visual performance of the generated face images from SF2F and voice2face. We compare SF2F and voice2face trained on HQ-VoxCeleb, to justify the performance boost brought by our model design. We also provide results generated by voice2face pre-trained and released by Wen \textit{et al.} \cite{v2f_github}, to prove the improvement brought by HQ-VoxCeleb. All of the images, from all three of the models, are reconstructed based on speech ranging from 4-7 seconds. As shown in Fig. \ref{fig:qualitative}, faces reconstructed by SF2F contain obviously preserves more accurate and meaningful facial features. The pose, expression, and lighting over the faces from SF2F are generally more stable and consistent. More qualitative results of 128 $\times$ 128 face image generation are available in Appendix \ref{section:qua_result}.
%

\noindent\textbf{Ablation study.} Ablation experiments are discussed in Appendix \ref{sec:abl_study}.

\section{Conclusion and Future Work}\label{sec:conclusion}

In this paper, we present \emph{Speech Fusion to Face} (SF2F), a new strategy of building generative models converting speech into vivid face images of the target individual. We demonstrate the huge performance boost, on both face similarity and image fidelity, brought by the enhancement of training data quality and the new vocal embedding fusion strategy. In the future, we will explore the following research directions for further performance improvement of \emph{speech2face} system. Firstly, we plan to introduce more accurate vocal embedding methods, in terms of the capability to distinguish different speakers. Secondly, we will evaluate the possibilities of hierarchical attention mechanisms, in order to link the vocal features to the corresponding facial features on the correct layers.

\section*{Broader Impact}


\textit{Speech Fusion to Face} (SF2F) makes the following positive impacts to the society: (1) with the help of SF2F, law enforcement departments can convert voiceprint evidence to face portraits, to facilitate tracking or identifying culprits; (2) SF2F can be used together with video processing techniques to create entertaining videos. On the other hand, the privacy protection of personal facial images will become more challenging. SF2F could be abused to infer facial images, even when the individual prefers to keep anonymous by using his voice only. When SF2F is applied for public security, inaccurate face reconstruction could wrongly identify innocent people as suspects, therefore, the police departments should only use SF2F as an auxiliary system in their crime fighting campaign. If SF2F is trained on a biased population with unbalanced geographical attributes, such as race, age, and gender, the performance of SF2F may worsen when used to predict facial images of minorities, consequently leading to potential fairness problem.

\small
\bibliographystyle{unsrt}

\newpage
\appendix

\section{Qualitative Results}\label{section:qua_result}
\begin{figure}[h]
    \centering
    \includegraphics[width=0.9\textwidth]{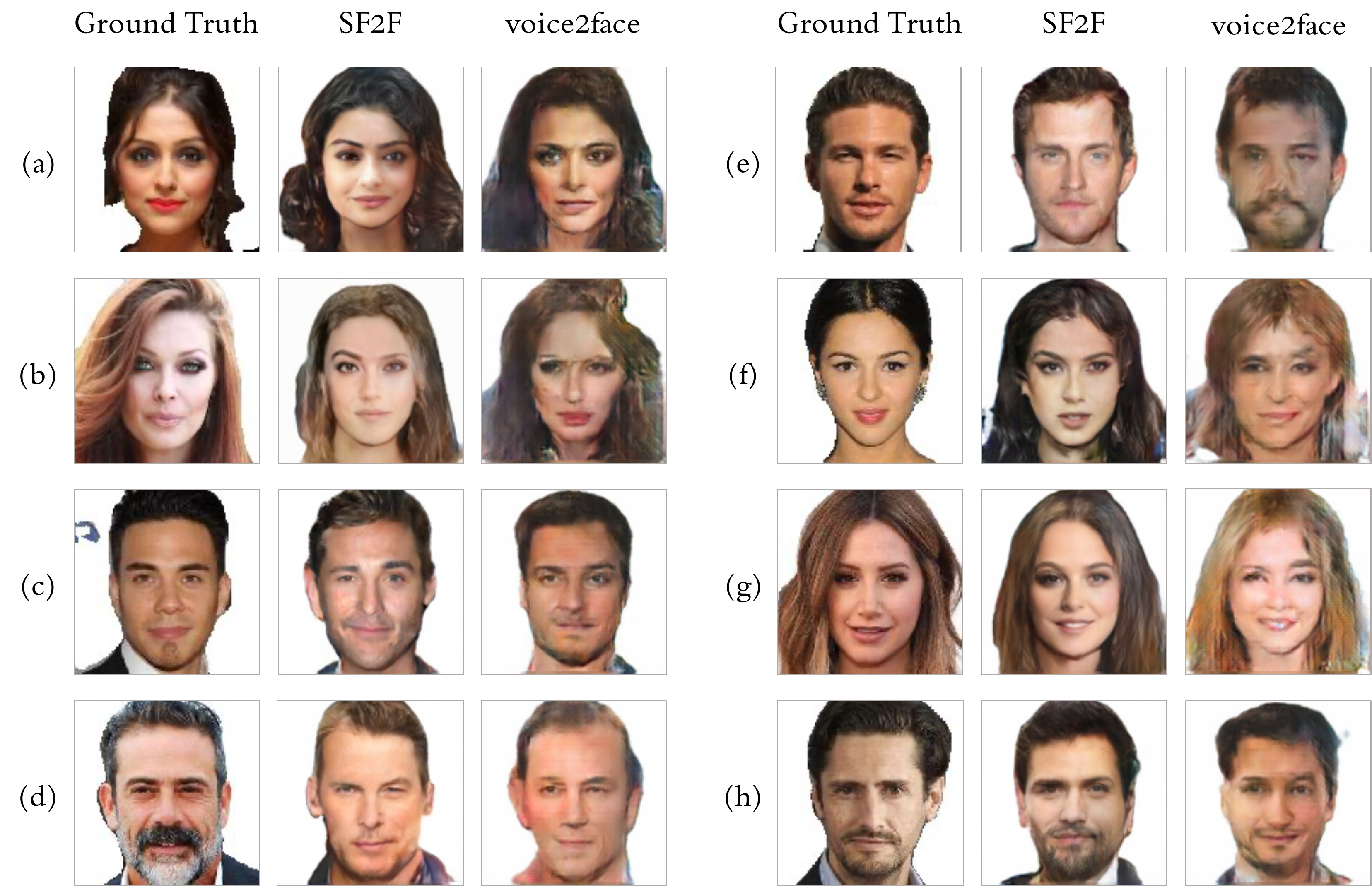}
    \vspace{0pt}
    \caption{Examples of 128 $\times$ 128 generated images using SF2F and voice2face}
    \label{fig:128_comparison}
    \vspace{-0pt}
\end{figure}

In the main paper, we compare $64 \times 64$ face images generated by SF2F and voice2face \cite{wen2019face}. In this section, we compare $128 \times 128$ face images reconstructed by both models. As the original voice2face \cite{wen2019face,v2f_github} is only trained to generate $64 \times 64$ images, we compare SF2F and voice2face trained on HQ-VoxCeleb dataset. To ensure fairness in the comparison, both models are trained until convergence, and checkpoints with the best L1 similarity are used for inference. As shown in Fig. \ref{fig:128_comparison}, although voice2face can capture attributes such as gender, SF2F generates images with much more accurate facial features as well as face contours. The pose, expression and lighting condition over the faces from SF2F are generally more stable and consistent than the face images from voice2face. In group (f), for example, SF2F predicts a face almost identical to the groundtruth, when the respective output from voice2face is hardly recognizable. This verifies our model enables more accurate information flow from the speech domain to the face domain.  

\section{Data Quality Enhancement}\label{section:dataset_apd}

\subsection{HQ-VoxCeleb Dataset}\label{sec:hq-vox}

\begin{figure}[t]
    \centering
    \includegraphics[width=0.8\textwidth]{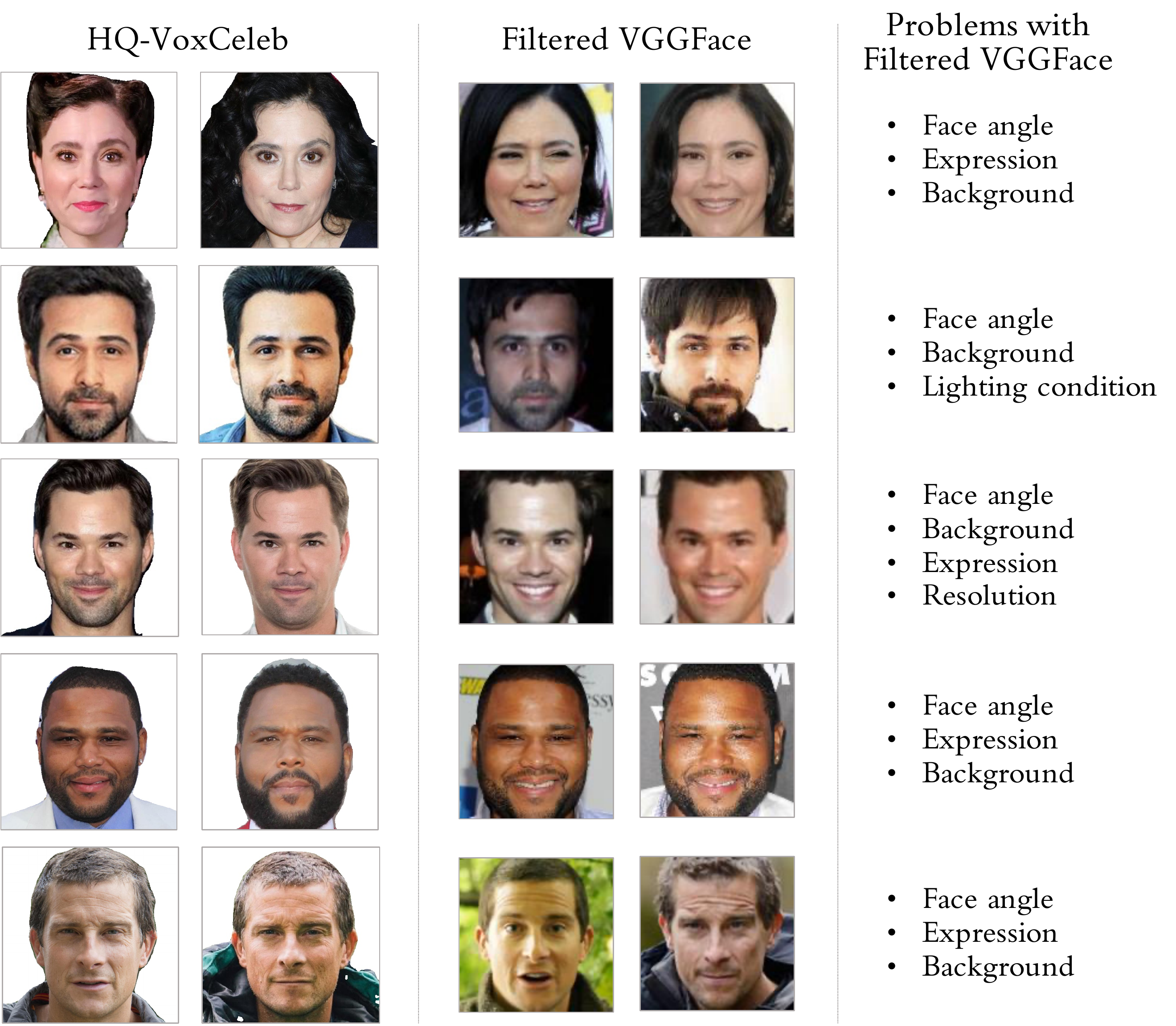}
    \vspace{0pt}
    \caption{The quality variance of the original manually filtered VGGFace dataset (on the right of the figure) \cite{parkhi2015deep} diverts the efforts of face generator model to the normalization of the faces. By filtering and correcting the face database (on the left), the computer vision models are expected to focus more on the construction of mapping between vocal features and physical facial features.}
    \label{fig:quality_comparison_apd}
    \vspace{-10pt}
\end{figure}

As an overview of the HQ-VoxCeleb dataset is presented in the main paper, in this section, we elaborate on the standards and process in our data enhancement scheme. As demonstrated in Fig. \ref{fig:quality_comparison_apd}, the poor quality of training dataset for \emph{speech2face} is one of the major factors hindering the improvement of \emph{speech2face} performance. To eliminate the negative impact of the training data itself, we carefully design and build a new high-quality face database on top of VoxCeleb dataset, such that face images associated with the celebrities are all at reasonable quality.

To fulfill the vision of data quality, we set a number of general guidelines over the face images as the underlying measurement on \emph{quality}, as follows:

\begin{itemize}
    \item[$\bullet$] \textbf{Face angle} between human's facial plane and the photo imaging plane is no larger than 5$^{\circ}$;
    \item[$\bullet$] \textbf{Lighting condition} on the face is generally uniform, without any noticeable shadow and sharp illumination change;
    \item[$\bullet$] \textbf{Expression} on the human faces is generally neutral, while minor smile expression is also acceptable;
    \item[$\bullet$] \textbf{Background} does not contain irrelevant information in the image, and completely void background in white is preferred.
\end{itemize}

To fully meet the standards as listed above, we adopt the following methodology to build the enhanced \emph{HQ-VoxCeleb} dataset for our \emph{speech2face} model training.

\noindent\textbf{Data Collection.} We collect 300 in-the-wild images for each of the 7,363 individuals in VoxCeleb dataset, by crawling images of the celebrities on the Internet. The visual qualities of the retrieved images are highly diverse. The resolution of the images, for example, ranges from 47$\times$59 to 6245$\times$8093 in pixels. Moreover, some of the images cover the full body of the celebrity of interest, while other images only include the face of the target individual. It is, therefore, necessary to apply pre-processing and filtering operations to ensure 1) the variance of the image quality is reasonably small; and 2) all the images are centered at the face of the target individuals.

\noindent\textbf{Machine Filtering.} To filter out unqualified images from the massive in-the-wild images, we deploy an automated filtering module, together with a suite of concrete selection rules, to eliminate images at poor quality, before dispatching the images for human filtering. In the filtering module, the algorithm first detects the landmarks from the raw face images. Based on the output landmarks of the faces, the algorithm identifies poorly posed faces, if the landmarks from left/right sides cannot be well aligned. Low-resolution face images, with distance between pupils covers fewer than 30 pixels, are also removed. Finally, a pre-trained CNN classifier \cite{arriaga2017real} is deployed to infer the emotion of the individual in the image, such that faces not recognized as "neutral'' emotion are also removed from the dataset. 

\noindent\textbf{Image Processing.} Given the face images passing the first round machine-based filtering, we apply a two-step image processing, namely \emph{face alignment} and \emph{image segmentation} in the second round of filtering. In the first step of \emph{face alignment}, the images are rotated and cropped to make sure both the pupils of the faces in all these images are always at the same coordinates. In the second step of \emph{image segmentation}, we apply a pyramid scene parsing network \cite{zhao2017pyramid} pre-trained on Pascal VOC 2012 dataset \cite{pascal-voc-2012} to split the target individuals from the background in the images. The background is then refilled with white pixels. Note that the second step is helpful because irrelevant noise in the background may potentially confuse the generation model.

\noindent\textbf{Human Filtering.} To guarantee the final quality of the images in \emph{HQ-VoxCeleb} dataset, human workers are employed to select 3 to 7 images at best qualities for each celebrity. Only celebrities with at least 3 qualified face images are kept in the final dataset.

\subsection{Comparison with Existing Datasets}

In the main paper, we summarize the statistics of the result dataset after the adoption of the processing steps above. In this section, we compare HQ-VoxCeleb with existing audiovisual datasets \cite{horiguchi2018face,kim2018learning, ephrat2018looking,nagrani2017voxceleb,chung2018voxceleb2} in terms of data quality and its impact to model training, in order to justify the contribution of HQ-VoxCeleb. Table \ref{tab:datasets} shows the attributes of existing audiovisual datasets. 

Existing datasets, including VoxCeleb \cite{chung2018voxceleb2,nagrani2017voxceleb} and AVSpeech \cite{ephrat2018looking}, contain a massive number of pairwise data of human speech and face images. However, the existing datasets are constructed by cropping face images and speech audio from in-the-wild online data, and the face images thus vary hugely in pose, lighting, and emotion, which makes the existing datasets unfit for end-to-end learning of speech-to-face algorithms. To the highest image quality of existing datasets, Wen \textit{et al.} \cite{wen2019face} use the intersection of the filtered VGGFace \cite{horiguchi2018face} and VoxCeleb with the common identities. However, as shown in Fig. \ref{fig:quality_comparison_apd}, the filtered VGGFace cannot meet the quality standards defined in section \ref{sec:hq-vox}. Moreover, HQ-VoxCeleb has triple as many identities as filtered VGGFace as shown in Table \ref{tab:datasets}. In conclusion, the high quality and reasonable amount of data make HQ-VoxCeleb the most suitable dataset for \textit{speech2face} tasks.

\begin{table}[h]
  \begin{center}
    \caption{The attributes of HQ-VoxCeleb and existing audiovisual datasets}
    \label{tab:datasets}
    \begin{tabular}{c|c|c|c|c}
      \toprule
      \textbf{ Dataset } & \textbf{ \#Identity } & \textbf{ \#Image } & \textbf{ \#Speech } & \textbf{ Normalized Faces } \\
      \midrule
      { FVMatching \cite{horiguchi2018face} } & 1,078 & 118,553 & 131,815 & No \\
      { FVCeleb \cite{kim2018learning} } & 181 & 181 & 239 & No \\
      { AVSpeech \cite{ephrat2018looking} } & 100,000+ & - & - & No \\
      { VoxCeleb \cite{nagrani2017voxceleb,chung2018voxceleb2} } & 7,245 & - & 1,237,274 & No \\
      { Filtered VGGFace \cite{parkhi2015deep,wen2019face} } & 1,225 & 139,572 & 149,354 & No \\
      { \textbf{HQ-VoxCeleb} } & 3,638 & 8,028 & 609,700 & \textbf{Yes} \\
      \bottomrule
    \end{tabular}
  \end{center}
\end{table}

\section{Additional Experimental Results}

\subsection{Ablation Study}\label{sec:abl_study}

In this section, we focus on the effectiveness evaluation over the model components, loss functions, and dataset used in our SF2F approach. 
The ablated models are trained to generate $64 \times 64$ images, with all the results are summarized in Table \ref{tab:ablation_study}.
%
When removing any of the four loss functions in $\{L_1, L_G, L_C, L_P\}$, the performance of SF2F drops accordingly. This shows it is necessary to include all these components to ensure the learning procedure of the model is well balanced.
1D-CNN encoder is used in the baseline approach \cite{wen2019face}, while SF2F employs 1D-Inception encoder instead. We report the performance of SF2F by substituting our encoder with 1D-CNN, referred to as \emph{Baseline Encoder} in Table \ref{tab:ablation_study}. This replacement causes a significant drop of the performance on all metrics. Similarly, by adopting the deconvolution-based decoder used in \cite{wen2019face,Oh_2019_CVPR} instead of the upsampling-and-convolution-based decoder in SF2F, referred as \emph{Baseline Decoder} in Table \ref{tab:ablation_study}, we observe a slight yet consistent performance drop.
%
The impact of data quality is also evaluated, by training the SF2F model with the manually filtered version of VGGFace dataset \cite{parkhi2015deep} instead of HQ-VoxCeleb, by including overlap celebrity individuals included in both datasets. Poor data quality obviously leads to a huge performance plunge, which further justifies the importance of training data quality enhancement.

\begin{table}[h]
  \setlength{\tabcolsep}{3pt}
  \vspace{-0mm}
  \caption{Ablation Study on HQ-VoxCeleb dataset}
  \small
  \begin{center}
  \begin{tabular}{c|cc|cccc|c}
    \toprule
    \multicolumn{1}{c}{ \textbf{Setting} } & \multicolumn{2}{|c}{\textbf{ Similarity }} & \multicolumn{4}{|c}{\textbf{ Retrieval }} & \multicolumn{1}{|c}{\textbf{ Quality }} \\ 
    {Ablation} & {cosine} & {L1} & {R@1} & {R@2} & {R@5} & {R@10} & {VFS} \\    
    \midrule
    {w/o $L_1$} & {0.310} & {21.74} & {3.47} & {7.26} & {18.36} & {31.56} & {16.84} \\
    {w/o $L_G$} & {0.298} & {20.97} & {3.23} & {8.70} & {16.73} & {32.36} & {18.51} \\
    {w/o $L_C$} & {0.310} & {18.13} & {5.24} & {8.45} & {17.30} & {34.28} & {19.16} \\
    {w/o $L_P$} & {0.236} & {18.22} & {2.73} & {5.23} & {15.39} & {26.37} & {18.44} \\
    {Baseline Encoder} & {0.309} & {18.47} & {5.54} & {9.26} & {19.18} & {34.75} & {18.89} \\
    {Baseline Decoder} & {0.311} & {18.52} & {4.32} & {8.56} & {19.47} & {34.25} & {18.46} \\
    {LQ. Dataset} & {0.267} & {19.94} & {1.96} & {4.21} & {10.67} & {21.34} & {16.73} \\
    \midrule
    {Full Model} & {0.317} & {17.91} & {5.75} & {9.32} & {20.36} & {36.65} & {19.49} \\ 

    \bottomrule
  \end{tabular}
  \end{center} 
  \vspace{-0mm}
  \label{tab:ablation_study}
\end{table}

\subsection{Performance on Filtered VGGFace}\label{sec:fvg_perf}

We also compare the performance of SF2F and voice2face \cite{wen2019face} on the filtered VGGFace dataset \cite{parkhi2015deep}, to evaluate how SF2F functions under less controlled conditions. To make a fair comparison, we train both models with filtered VGGFace dataset. Both voice2face and our SF2F are trained to generate images with 64$\times$64 pixels, and evaluated with 1.25 seconds and 5 seconds of human speech clips. As is presented in Table \ref{table:64_performace_comparison_fvg}, SF2F outperforms voice2face by a large margin on all metrics. By deploying fuser in SF2F, the maximal recall@10 reaches 21.34\%,  significantly outperforming voice2face.

\begin{table}[h]
  \setlength{\tabcolsep}{3pt}
  \caption{Performance on filtered VGGFace dataset in feature similarity, retrieval recall@K, and VGGFace Score}
  \small
  \begin{center}
  \begin{tabular}{cc|cc|cccc|c}
    \toprule
    \multicolumn{2}{c}{ \textbf{Setting} } & \multicolumn{2}{|c}{\textbf{ Similarity }} & \multicolumn{4}{|c}{\textbf{ Retrieval }} & \multicolumn{1}{|c}{\textbf{ Quality }} \\ 
    {Method} & {Len.} & {cosine} & {L1} & {R@1} & {R@2} & {R@5} & {R@10} & {VFS} \\
    \midrule
    {random} & {-} & {-} & {-} & {1.00} & {2.00} & {5.00} & {10.00} & {-}\\
    \midrule
    {voice2face} & {1.25 sec} & {0.164} & {26.45} & {1.46} & {3.12} & {7.25} & {20.71} & {12.94 $\pm$ 0.57} \\
    {SF2F (no fuser)} & {1.25 sec} & \textbf{0.251} & \textbf{21.23} & \textbf{1.81} & \textbf{4.05} & \textbf{9.77} & \textbf{18.56} & \textbf{16.62 $\pm$ 0.44} \\
       \midrule
   	{voice2face} & {5 sec} & {0.165} & {26.42} & {1.45} & {3.17} & {7.89} & {15.43} & {12.72 $\pm$ 0.73} \\
   	{SF2F (no fuser)} & {5 sec} & {0.248} & {21.46} & {1.77} & {3.98} & {9.75} & {18.36} & {16.49 $\pm$ 0.83} \\
   	{SF2F} & {5 sec} & \textbf{0.267} & \textbf{19.94} & \textbf{1.96} & \textbf{4.21} & \textbf{10.67} & \textbf{21.34} & \textbf{16.73 $\pm$ 0.67} \\
    \bottomrule
  \end{tabular}
  \end{center} 
  \vspace{-3mm}
  \label{table:64_performace_comparison_fvg}
\end{table}

\section{Model Details}
\label{sec:model}

\subsection{Voice Encoder}
We use a 1D-CNN composed of several 1D Inception modules \cite{szegedy2015going} to process mel-spectrogram. The voice encoder module converts each short input speech segment into a predicted facial feature embedding. The architecture of the 1D Inception module and the voice encoder is summarized in Table \ref{tab:voice_encoder}. The Inception module models various ranges of short-term mel-spectrogram dependency, and enables more accurate information flow from voice domain to image domain, compared to plain single-kernel-size CNN used in \cite{wen2019face}. 

\begin{table}[h]
  \begin{center}
    \caption{The detailed structure of the 1D Inception module and voice encoder. 
    In the descriptions, Conv $x$/$y$ denotes 1D convolution with kernel size of $x$ and stride length of $y$.}
    \label{tab:voice_encoder}
    \vspace{-5pt}
    \parbox{.58\linewidth}{
     \centering
     \small
     \begin{tabular}{c|c|c}
       \toprule
       \multicolumn{3}{c}{\textbf{ 1D Inception Module }} \\
       \midrule
       \textbf{ Component } & \textbf{ Activation } & \textbf{ Dimension } \\
       \midrule
       Input & - & { $d_{i} \times t_{i}$ } \\
       { Conv 2/2 } & { BN + ReLU } & { $d \times t_{o}$ } \\
       { Conv 3/2 } & { BN + ReLU } & { $d \times t_{o}$ } \\
       { Conv 5/2 } & { BN + ReLU } & { $d \times t_{o}$ } \\
       { Conv 7/2 } & { BN + ReLU } & { $d \times t_{o}$ } \\
       { Concat. } & - & { $4 d \times t_{o}$ } \\
       \bottomrule
    \end{tabular}
    }
    \hfill
    \parbox{.4\linewidth}{
     \hspace{-0.5cm}
     \centering
     \begin{tabular}{c|c}
       \toprule
       \multicolumn{2}{c}{\textbf{ Voice Encoder }} \\
       \midrule
       \textbf{ Layer } & \textbf{ Dimension } \\
       \midrule
       { Input } &  { $40 \times t_{0}$ } \\
       { Inception 1 } & { $256 \times t_{1}$ } \\
       { Inception 2 } & { $384 \times t_{2}$ } \\
       { Inception 3 } & { $576 \times t_{3}$ } \\
       { Inception 4 } & { $864 \times t_{4}$ } \\
       { Inception 5 } & { $512 \times t_{5}$ } \\
       { Time AvePool } & { $512 \times 1$ } \\
      \bottomrule
    \end{tabular}
    }
  \end{center}
  \vspace{-0pt}
\end{table}

\subsection{Face Decoder}

\subsubsection{Single-resolution Decoder}
\label{subsubsec:single-reso}

The face decoder reconstructs the target individual's face image based on the embeddings extracted from the individual's speech segments. The architectures of the upsampling block (UpBlock) and the face decoder are summarized in Table \ref{tab:face_decoder}. We show the structure of the face decoder generating 64$\times$64 images, and the face decoder generating 128$\times$128 images can be built by adding \emph{UpBlock 7} after \emph{UpBlock 6}. Our empirical evaluations prove that our decoder based on upsampling and convolution renders better performance in all the metrics compared to the deconvolution-based decoder employed in existing studies \cite{Oh_2019_CVPR} and \cite{wen2019face}. 

\begin{table}[h]
  \begin{center}
    \caption{The detailed structure of the upsampling block and face decoder. In the descriptions, Conv $x_{/y,z}$ denotes 2D convolution with kernel size of $x$, stride length of $y$, and padding size of $z$.}
    \label{tab:face_decoder}
    \vspace{0pt}
    \parbox{.5\linewidth}{
     \centering
     \small
     \begin{tabular}{c|c}
       \toprule
       \multicolumn{2}{c}{\textbf{ Upsampling Block }} \\
       \midrule
       \textbf{ Component } & \textbf{ Dimension } \\
       \midrule
       { Input } & { $d_{i} \times w_{i} \times w_{i}$ } \\
       { Upsampling } & { $d_{i}  \times 2w_{i} \times 2w_{i}$  } \\
       { Conv $3_{/2,1}$ } & { $d  \times 2w_{i} \times 2w_{i}$  } \\
       { ReLU } & { $d  \times 2w_{i} \times 2w_{i}$ } \\
       { Batch Norm } & { $d  \times 2w_{i} \times 2w_{i}$ } \\
       \bottomrule
    \end{tabular}
    }
    \hfill
    \parbox{.45\linewidth}{
     \hspace{-0.5cm}
     \centering
     \begin{tabular}{c|c}
       \toprule
       \multicolumn{2}{c}{\textbf{ Face Decoder }} \\
       \midrule
       \textbf{ Layer } & \textbf{ Dimension } \\
       \midrule
       { Input } &  { $512 \times 1 \times 1 $ } \\
       { UpBlock 1 } & { $1024 \times 2 \times 2 $ } \\
       { UpBlock 2 } & { $512 \times  4 \times 4 $ } \\
       { UpBlock 3 } & { $256 \times  8 \times 8 $ } \\
       { UpBlock 4 } & { $128 \times  16 \times 16 $ } \\
       { UpBlock 5 } & { $64 \times  32 \times 32 $ } \\
       { UpBlock 6 } & { $32 \times  64 \times 64 $ } \\
       { Conv $1_{/1,0}$ } & { $3 \times  64 \times 64 $ } \\
      \bottomrule
    \end{tabular}
    }
  \end{center}
  \vspace{0pt}
\end{table}

\subsubsection{Multi-resolution Decoder}

\begin{figure}[h]
\vspace{0pt}
\centering
\includegraphics[width=\columnwidth]{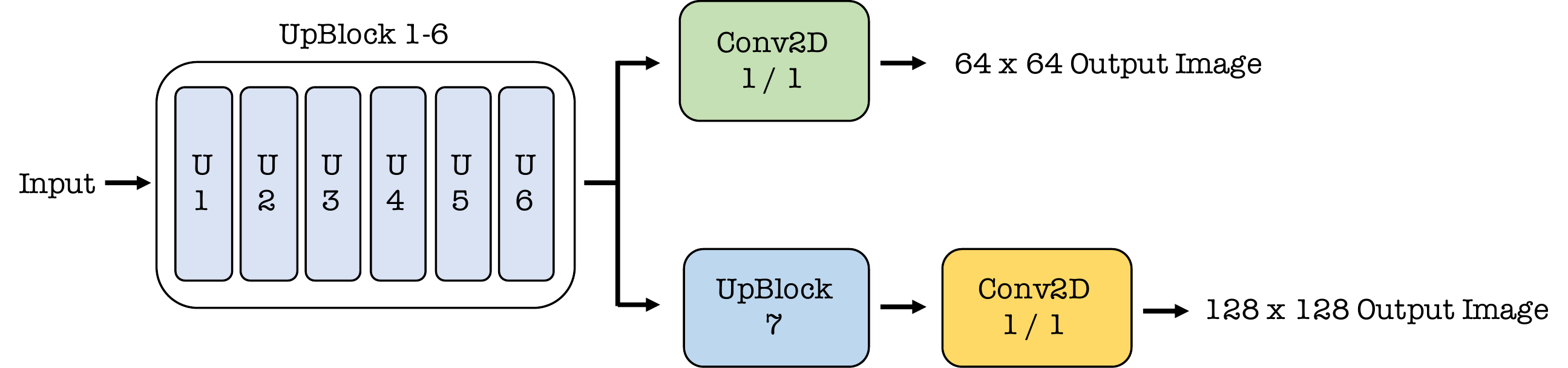}
\vspace{0pt}
\caption{Structure of the multi-resolution face decoder}
\label{fig:multi-reso}
\vspace{0pt}
\end{figure}

Inspired by \cite{zhang2018stackgan++}, the multi-resolution decoder is optimized to generate images at both low resolution and high resolution, which is shown in Fig. \ref{fig:multi-reso}. With the multi-resolution approach, the decoder learns to model the multiple target domain distributions of different scales, which helps to overcome the difficulty of generating high-resolution images.

\subsection{Discriminators}

The network structure of image discriminator $D_{real}$ and identity classifier $D_{id}$ is described in Table \ref{tab:discriminator}. Both networks are convolutional neural networks, followed by fully connected networks. In Table \ref{tab:discriminator}, we demonstrate the structure of discriminators for 64 $\times$ 64 images, and the discriminator of 128 $\times$ 128 images can be simply implemented by adding another convolution layer before average pooling layer. 

\begin{table}[h]
  \setlength{\tabcolsep}{3pt}
  \vspace{0mm}
  \caption{The structure of the discriminators for 64 $\times$ 64 images. In this table, Conv $x_{/y,z}$ denotes 2D convolution with kernel size of $x$, stride length of $y$, and padding size of $z$. $d_o = 2$ for binary image discriminator, and $d_o = N_{id}$ for identity classifier.}
  \begin{center} 
  \small
  \begin{tabular}{c|c|c}
       \toprule
       \multicolumn{3}{c}{\textbf{ Discriminator Network }} \\
       \midrule
       \textbf{ Component } & \textbf{ Activation } & \textbf{ Dimension } \\
       \midrule
       {Input} & - & { $3 \times  64 \times 64 $ } \\
       { Conv $4_{/2,1}$ } & { BN + ReLU } & { $64 \times 32 \times 32 $ } \\
       { Conv $4_{/2,1}$ } & { BN + ReLU } & { $128 \times 16 \times 16 $ } \\
       { Conv $4_{/2,1}$ } & { BN + ReLU } & { $256 \times 8 \times 8 $ } \\
       { Avg. Pooling } & { - } & { $256 \times 1 \times 1 $ } \\
       { Linear } & { ReLU } & { $1024$ } \\
       { Linear } & { Softmax } & { $d_o$ } \\
       \bottomrule
    \end{tabular}
  \end{center}
  \vspace{0pt}
  \label{tab:discriminator}
\end{table}

\subsection{Complexity Analysis of SF2F model}

Shown in Table \ref{tab:complexity} is the time and space complexity of each component of SF2F model, where $t$ stands for the length of the input audio and $w$ stands for the dimension of the output face image. The time complexity of convolution and attention is $\Omega(1)$ regardless of the input length. The space complexity of 1D convolution is $O(t)$ and the space complexity of self attention is $O(t^2)$.

\begin{table}[h]
  \begin{center}
    \caption{Complexity analysis}
    \label{tab:complexity}
    \begin{tabular}{c|c|c}
      \toprule
      \textbf{ Component } & \textbf{ Time Complexity } & \textbf{ Space Complexity } \\
      \midrule
      { Voice Encoder } & $\Omega(1)$ & $O(t)$ \\
      { Embedding Fuser } & $\Omega(1)$ & $O(t^2)$ \\
      { Face Decoder } & { $\Omega(1)$ } & $O(w^2)$ \\
      \bottomrule
    \end{tabular}
  \end{center}
\end{table}

\section{Implementation Details}

\subsection{Training}

We train all our SF2F models on 1 NVIDIA V100 GPU with 32 GB memory; SF2F is implemented with PyTorch \cite{paszke2019pytorch}. The encoder-decoder training takes 120,000 iterations, and the fuser training takes 360 iterations. The training of 64 $\times$ 64 models takes about 18 hours, and the training of 128 $\times$ 128 models takes about 32 hours. The model is trained only once in each experiment.

\subsection{Evaluation}

Evaluation is conducted on the same GPU machine. The details of the implementation are provided in this part of the section.

\paragraph{Ground Truth Embedding Matrix.} As mentioned in the main paper, the ground truth embedding matrix $U=\{u_1,u_2,\ldots,u_N\}$  extracted by FaceNet \cite{schroff2015facenet} is used for similarity metric and retrieval metric. Given that one identity is often associated with multiple, i.e., $K$, face images in both datasets, for an identity of index $n$, the ground truth embedding is computed by $
{u}_n = \sum^K_{j=1} u_{nj}/K$. The embedding $u_n$ is normalized as $u_n=\frac{u_n}{\max(\lVert u_n \rVert_2, \epsilon)}$, because the embeddings extracted by FaceNet are also L2 normalized. Building embedding matrix with all the image data helps remove variance in data. This makes the evaluation results fair and stable. 

\paragraph{Evaluation Runs.} In each experiment, we randomly crop 10 pieces of audio with desired length for each identity in the evaluation dataset. With the data above, each experiment is evaluated \emph{ten} times, the mean of each metric is calculated based on the outcomes of all these ten evaluation runs. We additionally report the variance of VGGFace Score.

\subsection{Hyperparameter}

The hyperparameters for SF2F's model training are listed in Table \ref{tab:hparam}. The detailed configuration of SF2F's network is available in Section \ref{sec:model} as well as the main paper.

\begin{table}[h]
  \begin{center}
    \caption{Hyperparameters of SF2F}
    \label{tab:hparam}
    \begin{tabular}{c|c}
      \toprule
      \textbf{ Hyperparameter } & \textbf{ Value } \\
      \midrule
      { Optimizer } & { Adam \cite{dpk2015adam} } \\
      { Optimizer $\beta_1$ } & 0.9 \\
      { Optimizer $\beta_2$ } & 0.98 \\
      { Learning Rate } & {5e-4} \\
      { Batch Size } & 256 \\
      { $\lambda_1$ } & 10 \\
      { $\lambda_2$ } & 1 \\
      { $\lambda_3$ } & 0.05 \\
      { $\lambda_4$ } & 100 \\
      { Fuser Dimension } & 512 \\
      \bottomrule
    \end{tabular}
  \end{center}
\end{table}

Hyperparameters are carefully tuned. As the training is time-consuming and the hyperparameter space is large, it is difficult to apply grid search directly. We adopt other methods for parameter tuning instead. 

We first train a SF2F model with $\lambda_1, \lambda_2, \lambda_3 = 1$ and $\lambda_4 = 0$, we adjust the parameters $\lambda_1$ and $\lambda_3$ and find that increasing the Image Reconstruction Loss weight $\lambda_1$ and decreasing the Auxiliary Classifier Loss weight $\lambda_3$ both improve the model performance. We adjust $\lambda_1$ and $\lambda_3$ gradually and find out the model achieves the best performance when $\lambda_1 = 10$ and $\lambda_3 = 0.05$ . Afterward, we apply Perceptual Loss to our model training with initial weight $\lambda_4 = 1$. We find increasing $\lambda_4$ improves SF2F's performance, which is because the original scale of Perceptual Loss is much smaller than the scale of the other three losses. We gradually increase $\lambda_4$ until we observe SF2F achieving the best performance when $\lambda_4 = 100$. Therefore, $\lambda_1, \lambda_2, \lambda_3, \lambda_4$ are set at 10, 1, 0.05 and 100, respectively.

Afterwards, we apply a grid search over learning rate and batch size. We test with different values, which are listed in Table \ref{tab:grid_search} with optimal values singled out in the last column.

\begin{table}[h]
  \begin{center}
    \caption{The values considered for grid search}
    \label{tab:grid_search}
    \begin{tabular}{c|c|c}
      \toprule
      \textbf{ Hyperparamter } & \textbf{ Values Considered } & \textbf{ Optimal Value }\\
      \midrule
      { Learning Rate } & 1e-4, 2e-4, 5e-4, 1e-3 & 5e-4 \\
      { Batch Size } & 64, 128, 256, 512 & 256 \\

      \bottomrule
    \end{tabular}
  \end{center}
\end{table}
 
The hyperparameters above are determined by tuning a $64 \times 64$ SF2F model with HQ-VoxCeleb dataset, and we find this hyperparameter configuration works well with $128 \times 128$ SF2F model. With this hyperparameter configuration, SF2F outperforms voice2face on filtered VGGFace dataset. Consequently, we opt to skip further  hyperparameter tuning on filtered VGGFace. The total cost of hyperparameter tuning is around 558 GPU hours on V100.

\clearpage

\end{document}